\documentclass[runningheads,orivec]{llncs}
\usepackage[T1]{fontenc}
\usepackage{cite}
\usepackage{amsmath,amssymb,amsfonts}
\usepackage{textcomp}
\usepackage{xcolor}
\usepackage{booktabs}
\usepackage{array}
\usepackage{makecell}
\usepackage{todonotes}

\graphicspath{{FIGURES/}}
\usepackage{graphicx}
\begin{document}
\title{Questionnaire-Guided Disaggregation of Energy Appliance Use for Domestic Smart Meter Data}
\titlerunning{Disaggregation of energy appliance use from domestic smart meters}
%
\author{Achal Nanjundamurthy \and
Rupam Misra \and
Suzanne Little\orcidID{0000-0003-3281-3471} \and
Alan F. Smeaton\orcidID{0000-0003-1028-8389}
}
\authorrunning{A. Nanjundamurthy et al.}
%
\institute{School of Computing, Dublin City University, Glasnevin, Dublin 9, Ireland \\
\email{alan.smeaton@dcu.ie}
}
\maketitle              
\begin{abstract}
Ireland's  smart metering programme records electricity use at 30-minute resolution, with smart meters installed in over 80\% of households as of late 2025. While this is useful for billing of smart, time-of-use tariffs, it is too coarse to capture use of domestic appliances.  We present a label-free disaggregation system that breaks usage data into 9 appliance categories by combining event detection for high-power loads with questionnaire-guided estimation. Our evaluation draws on four datasets: a calibration household with a commercial comparator, two public benchmarks (UK-DALE and REFIT) with per-appliance sub-metering, and a smart meter dataset of more than 4,800 years of use from 2,968 Irish consumers. Compared against two independently developed disaggregation systems our hybrid method combining analysis of usage data with questionnaire results, achieves the lowest whole-decomposition error on all buildings across the datasets, with better month-level  performance over 54 paired months ($p<0.001$, Holm-corrected). Our method provides useful advice on a household's energy consumption patterns and advice on how to reduce or shift usage on some appliances in order to reduce costs.

\end{abstract}
\section{Introduction}
Ireland's smart meter programme has placed interval-sampling smart meters in most residential premises nationwide. Each meter records a single number indicating how much electricity a household used during each 30-minute interval. In practice, the information that matters to a householder is usage disaggregation, for example knowing how much energy a tumble dryer or a washing machine used and whether shifting that appliance to a time of day with a cheaper tariff would be worthwhile. Recovering such insights is called  non-intrusive load monitoring (NILM), posed by Hart three decades ago \cite{ref1}.

The Irish setting for addressing NILM is more challenging than the situation most NILM research addresses.  Published results are trained on 1s to 6s sampled data in which individual appliance events are visible \cite{ref4}. Within a 30-minute sampling interval a kettle, shower pump or cooker hob  can start and finish or it can straddle two adjacent intervals. Another challenge is that supervised methods for NILM need sub-metered appliance annotations for training, that is, per-appliance measurements from sub-meters installed inside a home alongside the main meter, and this does not exist for Irish households. 

The  setting for our work offers an information source that most NILM research ignores namely the householder. A short questionnaire, asking what heats the water, whether there is an electric vehicle for charging and how many fridges are plugged in, takes minutes to answer but carries useful information and utility companies already routinely collect such information. Our work starts from the premise that at very low sampling rates, more is gained by asking how evidence from usage data and declared household contexts should be combined, and how much each contributes, than by trying to extract only from the usage data.

We present a hybrid system for appliance disaggregation that assigns each household to one of three processing paths based on answering a short questionnaire, detects usage loads and ranks each home against a peer group drawn from a dataset of usage  from 2,968 households. Our work is compared against two independently developed disaggregation systems, a signature-rule approach and an additive Factorial Hidden Markov model (FHMM). 


The remainder of the paper is organised as follows. Section~II reviews related work. Section~III describes the data, the system we developed and the evaluation design. Section~IV reports our evaluation and Section~V concludes the paper.

\section{Related Work}
In 1992, George Hart \cite{ref1} defined NILM as inferring appliance consumption from a single supply-point measurement using power-level signatures and finite-state appliance models. 
For appliance category detection, supervised deep learning currently produces the most accurate  results. Kelly and Knottenbelt trained denoising autoencoders and recurrent networks on 1 to 6 second UK-DALE data \cite{ref3}, and Zhang et al. improved on this with sequence-to-point convolutions \cite{ref4}. Both approaches need sub-metered training labels and a sampling rate at which appliance events are visible,  neither condition holds in our work. 

An alternative form of disaggregation treats the aggregate as generated by several hidden appliance processes running in parallel. Ghahramani and Jordan introduced  factorial HMM and its structured mean-field approximation \cite{ref8}. Kolter and Jaakkola specialised additive-FHMM inference to energy disaggregation \cite{ref9}. 
The FHMM approach evaluated against our work implements exactly this 
and later we show that this approach struggles at a 30-minute sampling rate.

A small body of work addresses sampling rates as low as ours. Zhao et al. \cite{ref12} treat very-low-rate disaggregation directly and supply the operational bounds we use for our always-on estimator. 
The common factor across these studies is that at 30 minutes the signal under-determines the decomposition, so auxiliary information has to supply some context. That  motivates our use of  questionnaire results and motivates our research  about how much each field in the questionnaire  is actually worth.

The questionnaire part of our work draws on literature from Conditional Demand Analysis (CDA), introduced by Parti and Parti \cite{ref19} and extended to end-use load profiles by Aigner et al. \cite{ref20}. 
We apply the same idea at the level of a single household where constants are attached to declared appliances instead of population regression coefficients. 
What NILM adds, and what our hybrid design exploits, is the meter signal that CDA never used. Beckel et al. \cite{ref21} showed household characteristics are classifiable from load shapes alone using Irish  data, which motivates our  peer-profile assignment. Kavousian et al. \cite{ref22} found whole-home magnitude and seasonal shape to be the most robust determinants of consumption, and Kang and Reiner \cite{ref18} establish weather as the dominant driver for 3,827 Irish households. Together these bring together two axes into which we group peer households. 

\section{Methodology}
\subsection{Data Sources and Evaluation}
Our evaluation draws on four data sources, summarised in Table~\ref{tab:datasources}.
A single  Irish  household is used for calibration with seven months of per-category percentage estimates from the householder's energy supplier app (Energia). This data was used for in-sample consistency checks only. Two  sub-metered benchmark datasets, UK-DALE buildings 1, 2, 4 and 5 \cite{ref6} and all 20  houses in the REFIT benchmark \cite{ref7}, were downsampled to 30-minute intervals
yielding 76 UK-DALE building-months and 366 REFIT house-months. Finally, we used anonymised usage data from a published dataset of 2,990 Irish smart meters \cite{smeaton-data}, of which we use data from 2,968 import-only meters representing over 4,800 years of usage. 

The benchmark ground truth inherits the limitations of instrumentation, and we measured those limits before relying on them. Named-category coverage (100 minus the mean unmetered share) is 62.2\% on UK-DALE building 1 but only 44.0\%, 42.2\% and 27.2\% on buildings 2, 4 and 5. A per-channel audit  shows that building 4 has no instrumented washing appliance and  building 5's fridge channel records an implausible 3.13kWh per month. Both benchmarks  derive their always-on baseline from overnight minima  which is an approximation. Two construction constants, a 15\% monthly cap on the derived always-on channel and a 4kW appliance-channel clip, are  guards chosen by the dataset constructors. Later we quantify the ranking's sensitivity to the cap and shows that the named-category results are  invariant to it. 

\begin{table*}[!htb]
\caption{Data sources -- benchmarks are downsampled to 30-minute intervals.
\label{tab:datasources}}
\centering
\begin{tabular}{@{}>{\raggedright\arraybackslash}p{2.9cm}>{\raggedright\arraybackslash}p{2.9cm}>{\raggedright\arraybackslash}p{2.9cm}>{\raggedright\arraybackslash}p{3.3cm}@{}}
\toprule
\textbf{Data source} & \textbf{Size} & \textbf{Ground truth} & \textbf{Role} \\
\midrule
Calibration household (Irish) & 1 home, 7 months & Energy supplier estimates from Energia app & In-sample consistency checks only \\
\addlinespace
UK-DALE \cite{ref6} & 4 buildings, 76 building-months & Sub-metered channels & Accuracy, blind and informed \\
\addlinespace
REFIT \cite{ref7} & 20 houses, 366 house-months & Sub-metered channels & Accuracy, blind and inventory-informed \\
\addlinespace
Irish smart meters (anonymised) \cite{smeaton-data} & 2,968 household smart meters, $\approx$4,880 house-years & None & Population-scale characteristics \\
\bottomrule
\end{tabular}
\end{table*}

\begin{figure*}[!htb]
\centering
\includegraphics[width=\textwidth]{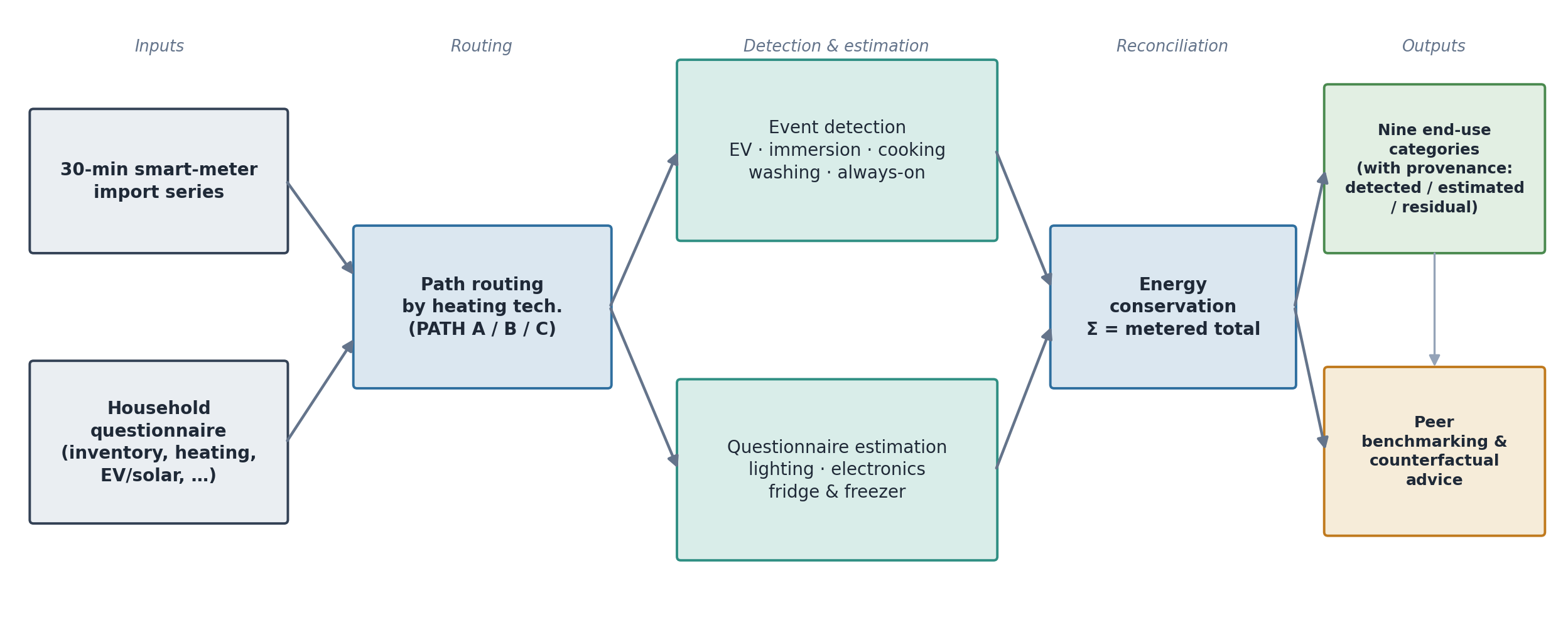}
\caption{Overview of the our pipeline.}
\label{fig:architecture}
\end{figure*}

\subsection{Output Classification and Path Routing}
Fig.~\ref{fig:architecture} gives an overview of our  pipeline. 30-minute meter data and a household questionnaire feed a path-routing stage. Appliances with exploitable structure are recovered by event detection while low-power loads are estimated from questionnaire responses. An energy-conservation stage then forces the 9 categories to sum to the metered total and the peer-benchmarking layer then ranks each home and generates counterfactual advice for the home owner. All approaches output usage across 9  categories (EV charging, water heating, cooking, washing, consumer electronics, lighting, fridge and freezer, always-on, and an unaccounted residual) through a shared category-mapping layer, so cross-approach comparison is exact. Each category carries a method label as detected, estimated or residual. Our hybrid approach  routes each household by its declared heating method.
PATH A covers full heat-pump homes where water heating is recovered as a monthly energy-balance residual after space heating. PATH B covers homes with a hot-water heat pump and non-electric space heating where water heating is the monthly residual after every other category is subtracted from gross consumption, with solar self-consumption added back for homes with PV. PATH C covers the Irish majority profile (gas or oil boiler with electric immersion backup), where water heating is detected directly as sustained 2.3kW to 3.8kW immersion heating events and the residual convention moves to the other category. 


\subsection{Detection, Estimation and Questionnaire Data}
Loads with exploitable half-hourly structure are detected directly. EV charging uses a two-part scheme: blocks of 5kW or more for wallbox charging, and 1.2kW to 5kW overnight blocks with a coefficient of variation of at most 0.15, followed by a machine-learning second stage. In that stage, Gaussian-mixture clustering of block features with isolation-forest anomaly rejection acts as a purely conservative false-positive filter. On a stratified sample of data from 200 smart meters  it removes 17 borderline detections (8.5\%) and never adds one. Removals were validated by manual inspection since no ground truth is available for these meters. Immersion water heating uses sustained-block signatures. Cooking uses events inside lunchtime and dinnertime windows above a power floor and is gated off for all-gas cooking profiles. Washing uses appliance-band events, and always-on uses overnight local minima clipped to a literature-determined band \cite{ref12}. Distributed low-power categories are estimated from questionnaire counts multiplied by constants derived from national statistics and standard conversion factors, with lighting scaled by number of bedrooms, house type and monthly dark hours. 

\label{sec:peer}

The questionnaire for households captures dwelling type, occupancy number, number of bedrooms, heating and hot-water methods, cooking fuel, appliance inventory, cold appliances and major assets (EV, solar). Default values are based on the most common Irish profile, so a blank questionnaire defaults into a blind mode, which we use for ablation. For peer comparison, households are grouped by annual-consumption tier and winter-to-summer  ratio. These are the two axes that the  literature on determinants finds most robust \cite{ref18,ref22} and  can be computed from  meter data alone. 

\subsection{Comparison Approaches}
We developed two approaches to smart meter disaggregation. 
The signature approach (\texttt{mprn\_v3}) follows Hart's signature-matching approach, using per-category power, duration and shape rules with fixed share fallbacks. It has no questionnaire pathway, so it represents signal-only analysis. The FHMM approach implements the additive factorial HMM with structured mean-field inference \cite{ref8,ref9}, Baum-Welch updates, time-of-day priors \cite{ref10} and a signal-aggregate-constraint regulariser \cite{ref11} followed by a Bayesian context-update stage that examines five questionnaire fields. 

The  metric we use for evaluation is  mean absolute deviation (MAD) between predicted and reference category shares as a  percentage, averaged over categories and months. 

\section{Performance Evaluation}
\subsection{Calibration-Household Consistency}
Table~\ref{tab:calib} compares our hybrid approach against a naive baseline that allocates consumption from questionnaire responses alone with no smart meter data processing, over seven months of the comparator used by the Energia app in our calibration household. The baseline uses the national-average EV constant which is about half this household's true charging volume.
Water heating is our pipeline's worst-performing category because PATH B computes it as a monthly residual that absorbs upstream error.
\begin{table}[!ht]
\caption{Per-category MAD for our pipeline vs. the commercial comparator from Energia on our calibration household over 7 months. Naive = questionnaire constants only, no smart meter data processing; pipeline = our full hybrid.
\label{tab:calib}}
\centering
\begin{tabular}{@{}lrr@{}}
\toprule
\textbf{Category} & \textbf{Naive} & \textbf{~~~Our Pipeline} \\
\midrule
EV charging & 8.73~\% & 7.70~\% \\
Water heating & 5.01~\% & 10.70~\% \\
Cooking & 4.07~\% & 2.17~\% \\
Consumer electronics & 2.00~\% & 1.77~\% \\
Lighting & 2.16~\% & 2.03~\% \\
Washing & 0.43~\% & 0.83~\% \\
Fridge and freezer & 4.57~\% & 2.31~\% \\
Always-on & 1.10~\% & 0.74~\% \\
Other & 2.00~\% & 0.29~\% \\
\midrule
Overall (9 categories) & 3.34~\% & 3.17~\% \\
Appliance suite (6 categories) & 2.39~\% & 1.64~\% \\
\bottomrule
\end{tabular}
\end{table}

Running the  pipeline without the questionnaire results included yields an overall MAD of 18.08\%, far worse than the questionnaire-only baseline at 3.34\%, while the full hybrid at 3.17\%. At this sampling resolution the questionnaire yields most of the accuracy on this household, and the data analysis contributes comparatively little. That observation motivated the value-of-information analysis in Section~\ref{sec:4-c}. Under the identical protocol, the signature approach records 4.38\% overall and the FHMM 6.20\%; the FHMM's Stage 2 context update improves the appliance suite while worsening the overall balance, an early sign of the prior-driven behaviour analysed below.

\subsection{Three Approaches to Appliance Disaggregation}
\label{sec:4-c}
Table~\ref{tab:ukdale} reports the overall MAD per approach on the four sub-metered UK-DALE buildings; the hybrid records the lowest error on every building in both modes.
\begin{table}[!t]
\caption{Overall MAD on UK-DALE, all 9 categories. Bold = best per building. All approaches are deterministic given their inputs, so no error bars  shown, dispersion across months is characterised by the paired tests of Table~\ref{tab:4}}
\label{tab:ukdale}
\centering
\footnotesize
\setlength{\tabcolsep}{4.5pt}
\begin{tabular}{@{}lrrrr@{}}
\toprule
\textbf{\makecell[l]{Building No.\\(months)}} & \textbf{Signature} & \textbf{FHMM} & \textbf{\makecell{Hybrid\\blind}} & \textbf{\makecell{Hybrid\\informed}} \\
\midrule
building1 (54) & 12.32~\% & 15.36~\% & 8.70~\% & \textbf{7.42~\%} \\
building2 (9) & 15.02~\% & 17.32~\% & \textbf{7.96~\%} & 9.95~\% \\
building4 (7) & 15.34~\% & 17.36~\% & \textbf{6.21~\%} & 8.89~\% \\
building5 (6) & 17.70~\% & 19.04~\% & \textbf{4.25~\%} & 5.76~\% \\
\bottomrule
\end{tabular}
\end{table}
Building 1 contributes 54 consecutive months of data, enough for paired testing. Table~\ref{tab:4} summarises Wilcoxon signed-rank tests on per-month MAD. All four comparisons survive Holm-Bonferroni correction at family-wise alpha 0.05 with large effect sizes. One caveat concerns independence as months within one household are seasonally autocorrelated, so the p-values are optimistic in absolute terms. For this reason we place more weight on the win counts than on marginal significance. Our hybrid approach produces lower error in each of the 54 months against the FHMM and in 53 of 54 against the signature approach. 

\begin{table}[!t]
\caption{Paired Wilcoxon tests, building 1, per-month MAD (54 months). Negative median favours the first approach
\label{tab:4}}
\centering
\footnotesize
\setlength{\tabcolsep}{4.5pt}
\begin{tabular}{@{}lrrrrr@{}}
\toprule
\textbf{Comparison} & \textbf{Wins} & \textbf{\makecell{Median\\(\%)}} & \textbf{W} & \textbf{p} & \textbf{r} \\
\midrule
Blind vs signature & 53/54 & $-3.62$ & 10 & $<0.001$ & 0.86 \\
Informed vs signature & 54/54 & $-4.81$ & 0 & $<0.001$ & 0.87 \\
Blind vs FHMM & 54/54 & $-6.77$ & 0 & $<0.001$ & 0.87 \\
Informed vs blind & 44/54 & $-1.22$ & 153 & $<0.001$ & 0.69 \\
\bottomrule
\end{tabular}
\end{table}

A leave-one-out and single-field ablation over seven questionnaire field groups on building 1 measures each field's marginal contribution (Table~\ref{tab:ablation}). Most of the questionnaire's value is in the appliance inventory, which on its own recovers 2.82\% when added to blind mode while removing it from the full profile costs 1.02\%. Occupancy has  zero effect and the negative deltas for bedrooms and house type are partly a coverage artefact as those fields feed the lighting estimate and UK-DALE's lamp-level lighting ground truth under-measures.

\begin{table}[!t]
\caption{Questionnaire field ablation for building 1
\label{tab:ablation}}
\centering
\begin{tabular}{@{}lrr@{}}
\toprule
\textbf{Field group} & \textbf{Leave-one-out} & \textbf{~~~~Single field vs. blind} \\
\midrule
Appliance inventory & $+1.02$~\% & $-2.82$~\% \\
Occupancy & 0.00~\% & 0.00~\% \\
Heating / hot-water type & 0.00~\% & -- \\
House type & $-0.36$~\% & $+0.02$~\% \\
Fridges/freezers & $-0.47$~\% & $-0.34$~\% \\
Bedrooms & $-0.72$~\% & $+0.04$~\% \\
\bottomrule
\end{tabular}
\end{table}

We  investigated why  informed mode performs worse than blind on buildings 2, 4 and 5. We isolated and quantified the mechanisms involved. The first is benchmark coverage. The informed-minus-blind gap tracks named coverage: minus 1.28\%  at 62\% coverage, against plus 1.99\%, 2.68\% and 1.51\% at 44\%, 42\% and 27\%. Informed mode moves energy into named categories, and a benchmark that left most consumption unmetered penalises exactly such behaviour. The second reason is due to dead and implausible channels. Building 4's washing ground truth derives from no instrument, and building 5's fridge channel implies a 3.13kWh per month refrigerator. On the audited, physically plausible channels, informed wins every  month on both buildings: 54 of 54 on building 1 (washing plus fridge gap minus 6.58\%) and 9 of 9 on building 2 (minus 1.91\%). The third reason is that declared ownership says little about how intensively an appliance is used. Where usage is typical, the informed predictions are close (building 1 washing 41.8 predicted vs. 45.2 measured kWh per month, fridge 28.5 vs. 22.6). Where usage is atypical, the constants miss in either direction: building 4's small fridge is overshot 2.1-fold (20.2 vs. 9.7), building 5's near-unused washing machine up to 9.0-fold (30.2 vs. 3.4), and building 2's heavily used machine is undershot (9.9 vs. 27.4). A questionnaire can only report what a household owns but how those appliances are actually used is invisible to it.

Examining the principal detector constants on building 1 shows a single dominant constant, the always-on clip ceiling,  from 0.030 to 0.110 kW, moves overall MAD across a 2.09\% range, with an interior minimum at 0.040 to 0.050 kW that replicates on all four buildings. The immersion band, cooking floor and washing band sit on plateaus but we are wary of reading too much into this.

\begin{figure}[!t]
\centering
\includegraphics[width=\columnwidth]{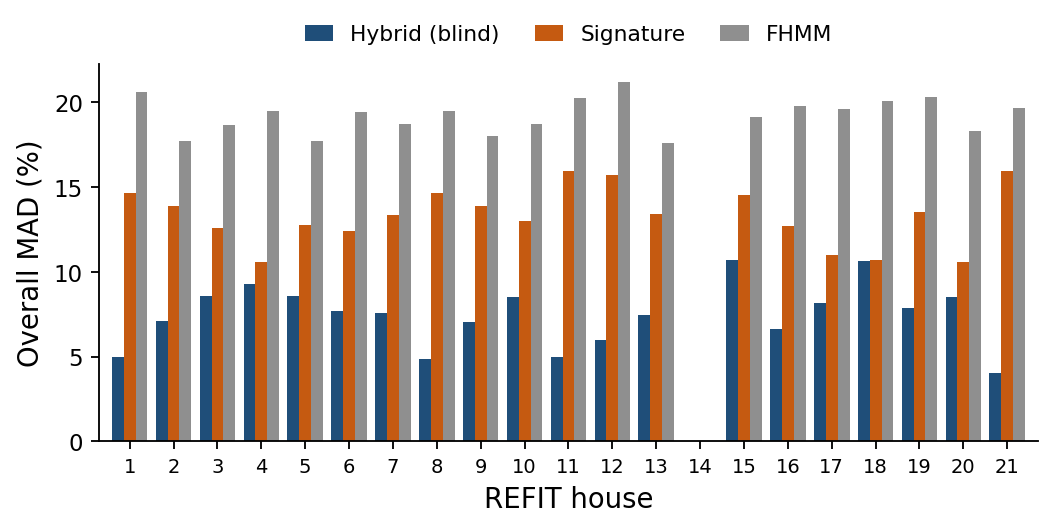}
\caption{Overall MAD per approach on each of 20 REFIT houses in blind mode. House 14 is absent because it is not included in the CLEAN REFIT release. Our hybrid approach yields the lowest error on every house.
\label{fig:refitmad}}
\end{figure}

The same approaches were applied to each of the 20 REFIT houses (366 house-months) in blind mode since REFIT provides no questionnaire data. Our hybrid approach records the lowest overall MAD on each of the 20 houses (Fig.~\ref{fig:refitmad}) with a mean of 7.46\% vs. 13.30\% for the signature approach and 19.23\% for  FHMM, with every pairwise Wilcoxon at W = 0 and $p<0.001$. The narrowest margin is on house 18, where the hybrid is best by only 0.027\%. A coverage-stratified analysis confirms the first mechanism from Section~\ref{sec:4-c} from the opposite direction. The hybrid's margin narrows as named coverage rises (5.69\% mean on the low-coverage tertile, 8.00\% on mid, and 8.76\% high), while the signature approach improves from 14.92\% to 11.64\%. This pattern makes sense for a residual-conservative allocator as the less of a house's consumption the benchmark actually metered, the better it performs.

\subsection{Validity of Results}

The nine-category MAD metric combines categories whose ground truths differ in that only 4 are directly sub-metered on the benchmarks we used (cooking, washing, consumer electronics, fridge), 2 are  absent (EV and electric water heating) and the remaining 3 are always-on which comes from an overnight-minimum heuristic, `other' is residual bookkeeping, and `lighting' is partially instrumented or not instrumented at all. Table~\ref{tab:subset} recomputes results for each approach on  category subsets.

\begin{table}[!htb]
\caption{MAD by category subset. Named-4 = cooking, washing, electronics, fridge. REFIT values are 20-house means. 
\label{tab:subset}}
\centering
\footnotesize
\setlength{\tabcolsep}{6pt}
\begin{tabular}{@{}lcc@{}}
\toprule
\textbf{approach} & \textbf{Full 9} & \textbf{Named-4} \\
\midrule
\multicolumn{3}{@{}l}{\textit{UK-DALE building1 (blind):}} \\
\quad Signature & 12.32~\% & 8.46~\% \\
\quad FHMM & 15.36~\% & 6.23~\% \\
\quad Hybrid blind & 8.70~\% & 9.63~\% \\
\quad Hybrid informed & 7.42~\% & 5.37~\% \\
\addlinespace
\multicolumn{3}{@{}l}{\textit{REFIT, 20 houses (blind):}} \\
\quad Signature & 13.30~\% & 6.40~\% \\
\quad FHMM & 19.23~\% & 11.47~\% \\
\quad Hybrid blind & 7.46~\% & 7.25~\% \\
\quad Hybrid informed (inventory) & 7.72~\% & 6.42~\% \\
\bottomrule
\end{tabular}
\end{table}

A number of  findings follow from Table~\ref{tab:subset}. At the whole decomposition level our hybrid approach is best on all 24 buildings, and much of our margin comes from not assigning energy to absent loads.  FHMM attributes 29.4\% of monthly error to a non-existent EV and 31.6\% to non-existent electric water heating in building 1, while the hybrid correctly predicts near zero for both. At the named-appliance level in blind mode however, our hybrid is not the best approach. FHMM leads the sub-metered categories on three of four UK-DALE buildings, and the signature approach leads on 13 of 20 REFIT houses (6.40\% against 7.25\%). 
Taken together, the approach ranking and the value-of-information analysis point to the same conclusion: at 30-minute resolution, appliance-level accuracy comes from declared context via the questionnaire. 

Running our pipeline over the 2,968-home dataset grounds the benchmarking layer. The always-on share of the combined always-on plus fridge pool has a median of 69.3\% with a near-degenerate interquartile range. 
The winter\/summer ratio across this data partitions into 29.9\%, 58.1\% and 12.0\% bands at cuts of 1.10 and 1.50. The cut values were chosen because they produce stable peer groups, and we make no claim that they correspond to natural breaks in the distribution. EV charging amounting to more than 5\% of household consumption is detected in 283 of the 2,968 meters (9.5\%).

\begin{figure*}[!t]
\centering
\includegraphics[width=\textwidth]{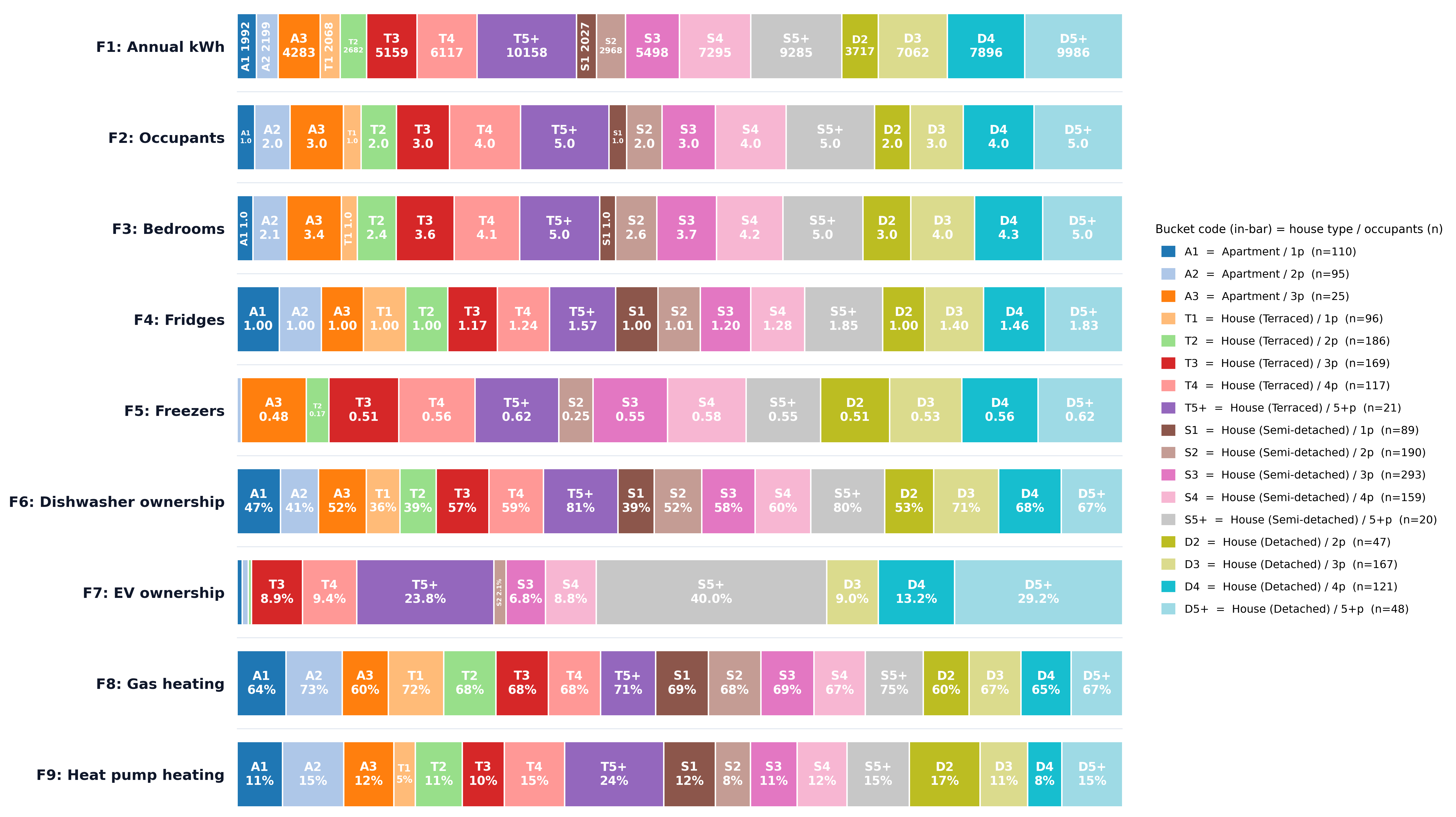}
\caption{Feature distributions of the synthetic household profiles (n = 1,962) used to build peer groups. Each row corresponds to a questionnaire-derived feature shown as a stacked bar across household buckets (dwelling type $\times$ occupancy). Each segment's width is that bucket's normalised share of its row total.}
\label{fig:featureprofile}
\end{figure*}

The peer household groups introduced in Section~\ref{sec:peer} are built from a set of synthetic household profiles, and Fig.~\ref{fig:featureprofile} shows their profiles grouped into buckets by dwelling type and occupancy. The numbers of each of the 9 questionnaire-derived features is shown across those buckets as ratios. Within each dwelling type, occupants, bedrooms and annual consumption increase with the number of occupants, larger dwelling types generally sit above smaller ones, and EV ownership concentrates in  larger, higher-occupancy buckets. Ownership of freezers and dishwashers are as one would expect of appliances that are not proportional to floor area. We repeat the caveat from earlier that these  are assigned from consumption-tier signals and Irish national statistics (Section~\ref{sec:peer}).

Applying the existing peer-benchmarking layer to every meter in the batch reveals usage characterises which could be provided as advice to the consumer on how they deviate from peers and ultimately where they could reduce or shift energy consumption. Table~\ref{tab:population} reports, per category, the \% of homes which are \texttt{high\_users}, the median gap to the peer group's own median in kWh and EUR per year (at an indicative EUR 0.35/kWh), and how often a synthetic-profile home's structural and attribute peer lenses disagree by two or more quartile bands on the same category. \texttt{always\_on} is excluded. 

\begin{table}[!t]
\caption{Population-scale advice characterisation, 2,968-meters. Gap = median gap-to-peer-median for above-median homes. Disagree = synthetic-profile subset (n=1,962) where structural and attribute panels place the same home $\geq$2 bands apart.
\label{tab:population}}
\centering
\footnotesize
\setlength{\tabcolsep}{4pt}
\begin{tabular}{@{}p{1.55cm}rrrr@{}}
\toprule
\textbf{Category} & \textbf{\makecell{High\\\%}} & \textbf{\makecell{Gap\\kWh/yr}} & \textbf{\makecell{Gap\\EUR/yr}} & \textbf{\makecell{Disagree\\\%}} \\
\midrule
EV charging & 20.0~\% & 147 & 51 & 7.7~\% \\
Water heating & 24.8~\% & 117 & 41 & 10.2~\% \\
Cooking & 24.9~\% & 115 & 40 & 4.0~\% \\
Consumer electronics & 24.4~\% & 25 & 9 & 30.3~\% \\
Lighting & 24.9~\% & 90 & 31 & 28.6~\% \\
Washing & 24.9~\% & 124 & 43 & 3.9~\% \\
Fridge and freezer & 24.7~\% & 38 & 13 & 30.5~\% \\
Other & 24.9~\% & 316 & 111 & 22.9~\% \\
\bottomrule
\end{tabular}
\end{table}

\section{Conclusions}
Our method for disaggregating Irish 30-minute smart-meter data without  training labels  processes it into a form a householder can inspect with a view to reducing of shifting their energy consumption. Our hybrid method yields the lowest whole-decomposition error on all 24 sub-metered buildings across two public benchmarks, with best month-level performance where paired testing is possible. 
The value of information from a user questionnaire's is real but narrowly concentrated, and depends on the household being typical. The occupancy value carries little importance, informed context helps precisely where ground truth can see it, on both benchmarks, and per-appliance constants fail for atypical users because ownership does not indicate actual usage.

We take two broad lessons from this. For utility companies, a three-item appliance inventory adds more accuracy at 30-minute sampling  than deeper signal analysis would and collecting context information from a small survey is cheaper than building a larger model. For researchers and householders, aggregate and appliance-level metrics can rank the same approaches in opposite orders on partially instrumented benchmarks. An evaluation that does not audit its own instrument and decompose its own metric will misattribute credit. 


\begin{credits}

\subsubsection{\discintname}
The authors have no competing interests to declare that are
relevant to the content of this article. 
\end{credits}
%
%
%
\bibliographystyle{splncs04}
\bibliography{Biblio}

\end{document}